\documentclass[11pt]{article}
\usepackage{acl2016}
\usepackage{times}
\usepackage{url}
\usepackage{latexsym}
\usepackage{slashbox}
\usepackage{tabulary}
\usepackage{tabularx,ragged2e}
\usepackage{arydshln}
\usepackage{lipsum}
\usepackage{enumitem}
\usepackage{amsmath}
\usepackage{pgfplots}
\usepackage{enumitem}
\pgfplotsset{compat=1.10}

\usepackage{array}
\newcolumntype{L}[1]{>{\raggedright\let\newline\\\arraybackslash\hspace{0pt}}m{#1}}
\newcolumntype{C}[1]{>{\centering\let\newline\\\arraybackslash\hspace{0pt}}m{#1}}
\newcolumntype{R}[1]{>{\raggedleft\let\newline\\\arraybackslash\hspace{0pt}}m{#1}}

\title{On the Contribution of Discourse Structure \\on Text Complexity Assessment}

\author{Elnaz Davoodi \\
        Concordia University\\
	    Department of Computer Science \\and Software Engineering\\
	    Montr{\'e}al, Qu{\'e}bec, Canada H3G 2W1\\
  {\tt e\_davoo@encs.concordia.ca} \\\And
        Leila Kosseim \\
        Concordia University\\
	    Department of Computer Science \\and Software Engineering\\
	    Montr{\'e}al, Qu{\'e}bec, Canada H3G 2W1\\
  {\tt kosseim@encs.concordia.ca } \\
  }

\date{}

\begin{document}
\maketitle
\begin{abstract}
This paper investigates the influence of discourse features on text complexity assessment. To do so, we created two data sets based on the Penn Discourse Treebank and the Simple English Wikipedia corpora and compared the influence of coherence, cohesion, surface, lexical and syntactic features to assess text complexity. 

Results show that with both data sets coherence features are more correlated to text complexity than the other types of features. In addition, feature selection revealed that with both data sets the top most discriminating feature is a coherence feature.
\end{abstract}

\section{Introduction}
\label{Introduction}

Measuring text complexity is a crucial step in automatic text simplification where various aspects of a text need to be simplified in order to make it more accessible \cite{siddharthan2014survey}. Despite much research on identifying and resolving lexical and syntactic complexity (e.g. \newcite{kauchak2013}, \newcite{rello2013}, \newcite{bott2012}, \newcite{carroll1998}, \newcite{barlacchi2013}, \newcite{vstajner2013}), discourse-level complexity remain understudied \cite{siddharthan2006,siddharthan2003}. Current approaches to text complexity assessment consider a text as a bag of words or a bag of syntactic constituents; which is not powerful enough to take into account deeper textual aspects such as flow of ideas, inconsistencies, etc. that can influence text complexity.


For example, according to \newcite{williams2003}, Example 1.a below is more complex than Example 1.b even though both sentences use exactly the same nouns and verbs.

 \medskip \textbf{Example 1.a.} \textit{Although many people find speed reading hard, if you practice reading, your skills will improve.}

\medskip \textbf{Example 1.b.} \textit{Many people find speed reading hard. But your skills will improve if you practice reading.} 
\\

Apart from the choice of words or the way these words form syntactically sound constituents, the way these constituents are linked to each other can influence its complexity. In other words, discourse information plays an important role in text complexity assessment.

The goal of this paper is to analyse the influence of discourse-level features for the task of automatic text complexity assessment and compare their influence to more traditional linguistic and surface features used for this task.



\section{Background}
\label{background}
A reader may find a text easy to read, cohesive, coherent, grammatically and lexically sound or on the other hand may find it complex, hard to follow, grammatically heavy or full of uncommon words. Focusing only on textual characteristics and ignoring the influence of the readers, \newcite{siddharthan2014survey} defines \textit{text complexity} as a metric to measure linguistic complexities at different levels of analysis: 1) lexical (e.g. the use of less frequent, uncommon and even obsolete words), 2) syntactic (e.g. the extortionate or improper use of passive sentences and embedded clauses), and 3) discourse (e.g. vague or weak connections between text segments).

Text complexity should be distinguished from \textit{text readability}. Whereas text complexity is reader-independent, text readability is reader-centric. According to \newcite{dale1949}, the readability of a text is defined by its complexity as well as characteristics of the readers, such as their background, education, expertise, level of interest in the material and external elements such as typographical features (e.g. text font size, highlights, etc.). It is crucial that a reader have access to a text with the appropriate readability level (e.g. \newcite{collins2014computational}, \newcite{williams2003}). An article which would be perceived as easy to read by a more educated or an expert reader may be hard to follow for a reader with a lower educational level. 

 Traditionally, the level of complexity of a text has mostly been correlated with surface features such as word length (the number of characters or number of syllables per word) or sentence length. One of the most well-known readability indexes, the Flesch-Kincaid index \cite{kincaid1975}, measures a text's complexity level and maps it to an educational level. Traditional complexity measures (e.g. \cite{chall1958,klare1963,zakaluk1988}) mostly consider a text as a bag of words or bag of sentences and rely on the complexity of a text's building blocks (e.g. words or phrases). This perspective does not take discourse properties into account. \newcite{webber2012discourse} define discourse using fours aspects: \textit{position of constitutes}, \textit{order}, \textit{context} and \textit{adjacency}. Such discourse information plays an important role in text complexity assessment. Traditional methods do not consider the flow of information in terms of word ordering, phrase adjacency and connection between text segments; all of which can make a text hard to follow, non-coherent and more complex.

More recently, some efforts have been made to improve text complexity assessment by considering richer linguistic features. For example, \newcite{schwarm2005} and \newcite{callan2007combining} used language models to predict readability level by using different language models (e.g. a language model for children using children's book, a language model for more advanced readers using scientific papers, etc.).

Discourse features can refer to text cohesion and coherence. Text cohesion refers to the grammatical and lexical links which connect linguistic entities together; whereas text coherence refers to the connection between ideas. Several theories have been developed to model both cohesion (e.g. centering theory \cite{grosz1995centering}) and coherence (e.g. Rhetorical Structure Theory \cite{mann1987}, DLTAG \cite{webber2004}). \newcite{pitler2008} examined a set of cohesion features based on an entity-based approach \cite{barzilay2008} and pointed out that these features were not significantly correlated with text complexity level. However to our knowledge, the influence of coherence on text complexity has not been studied.



\section{Complexity Assessment Model}
\label{methodology}

The goal of this study is to evaluate the influence of coherence features for text complexity assessment. To do so, we have considered various classes of linguistic features and build a pairwise classification model to compare the complexity of pairs of texts using each class of feature. For example, given the pair of sentences of Example 1.a and 1.b (see Section \ref{Introduction}), the classifier will indicate if 1.a is simpler or more complex than 1.b.

\subsection{Data Sets}

\label{data}

To perform the experiments, we created two different data sets using standard corpora. The first data set was created from the Penn Discourse Treebank (PDTB) \cite{Prasad08}; while, the other was created from the Simple English Wikipedia (SEW) corpus \cite{coster2011}. These two data sets are described below and summarized in Table \ref{Dataset}.

\subsubsection{The PDTB-based Data Set}
Since we aimed to analyze the contribution of different features, we needed a corpus with different complexity levels where features were already annotated or could automatically be tagged. Surface, lexical, syntactic and cohesion features can be easily extracted; however, coherence features are more difficult to extract. Standard resources typically used in computational complexity analysis such as the Simple English Wikipedia \cite{coster2011}, Common Core Appendix B\footnote{https://www.engageny.org} and Weebit \cite{vajjala2012} are not annotated with coherence information; hence these features would have to be induced automatically using a discourse parser (e.g. \newcite{LinNK14}, \newcite{LaaliDK15}).

\begin{table*}
\centering
\small
\begin{tabular}{||l|c|c||}
\hline 
\textbf{} & \textbf{PDTB-based Data Set}& \textbf{SEW-based Data Set}\\\hline\hline

\textbf{Source} & Penn Discourse& Simple English \\
& Treebank Corpus& Wikipedia Corpus\\ \hline

\textbf{\# of pairs of articles} & 378 & 1988\\\hline
\textbf{\# of positive pairs} & 194 & 944 \\ \hline
\textbf{\# of negative pairs} & 184 & 944 \\ \hline
\textbf{Discourse Annotation} & Manually Annotated & Extracted using \\
& & End-to-End parser \cite{LinNK14}\\ \hline

\end{tabular}
\caption{\small Summary of the two data sets.}
\label{Dataset}
\end{table*}

In order to have better quality discourse annotations, we used the data set generated by \newcite{pitler2009}. This data set contains 30 articles from the PDTB \cite{Prasad08} which are annotated manually with both complexity level and discourse information. The complexity level of the articles is indicated on a scale of 1.0 (easy) to 5.0 (difficult). Using this set of articles, we built a data set containing pairs of articles whose complexity levels differed by least $n$ points. In order to have a balanced data set, we set $n$ = 0.7. As a result, our data set consists of 378 instances with 194 positive instances (i.e. same complexity level where the difference between the complexity scores is smaller or equal to 0.7) and 184 negative instances (i.e. different complexity levels where the difference between complexity scores is larger than 0.7). Then, each pair of articles is represented as a feature vector where the value of each feature is the difference between the values of the corresponding feature in each article. For example, for a given pair of articles $<a_1, a_2>$, the corresponding feature vector will be:
\\
\begin{center}
\small
$V_{a_1,a_2}=<F_1^{a_1}-F_1^{a_2}, F_2^{a_1}-F_2^{a_2}, ..., F_{n}^{a_1}-F_{n}^{a_2} >$
\end{center}

where $V_{a_1,a_2}$ represents the feature vector of a given pair of articles $<a_1, a_2>$, $F_i^{a_1}$ corresponds to the value of the $i^{th}$ feature for article $a_1$ and $F_i^{a_2}$ corresponds to the value of the $i^{th}$ feature for article $a_2$ and $n$ is the total number of features (in our case $n = 14$ (see Section 3.2)).

Because the \newcite{pitler2009} data set is a subset of the PDTB, it is also annotated with discourse structure. The annotation framework of the PDTB is based on the DLTAG framework \cite{webber2004}. In this framework, 100 discourse markers (e.g. \textit{because}, \textit{since}, \textit{although}, etc.) are treated as predicates that take two arguments: Arg1 and Arg2, where Arg2 is the argument that contains the discourse marker. The PDTB annotates both explicit and implicit discourse relations. Explicit relations are explicitly signaled with a discourse marker. On the other hand implicit relations do not use an explicit discourse marker; however the reader still can infer the relation connecting the arguments. Example 2.a taken from \newcite{Prasad08} shows an explicit relation which is changed to an implicit one in Example 2.b by removing the discourse marker \textit{because}.

\medskip \textbf{Example 2.a.} \textit{If the light is red, stop \underline{because} otherwise you will get a ticket.}

\medskip \textbf{Example 2.b.} \textit{ If the light is red, stop. Otherwise you will get a ticket.} 
\\

In addition to labeling discourse relation realizations (i.e. explicit or implicit) and discourse markers (e.g. \textit{because}, \textit{since}, etc.), the PDTB also annotates the sense of each relation using three levels of granularity. At the top level, four classes of senses are used: \textsc{Temporal, Contingency, Comparison} and \textsc{Expansion}. Each class is expanded into 16 second level senses; themselves subdivided into 23 third-level senses. In our work, we considered the 16 relations at the second-level of the PDTB relation inventory\footnote{These are: Asynchronous, Synchronous, Cause, Pragmatic Cause, Condition, Pragmatic Condition, Contrast, Pragmatic Contrast, Concession, Pragmatic Concession, Conjunction, Instantiation, Restatement, Alternative, Exception, List.}.

\subsubsection{The SEW-based Data Set}
In order to validate our results, we created a larger data set but this time with induced discourse information. To do so, a subset of the Simple English Wikipedia (SEW) corpus \cite{coster2011} was randomly chosen to build pairs of articles. The SEW corpus contains two sections that are 1) article-aligned and 2) sentence-aligned. We used the article-aligned section which contains around 60K aligned pairs of regular and simple articles. Since this corpus is not manually annotated with discourse information, we used the End-to-End parser \cite{LinNK14} to annotate it. In total, we created 1988 pairs of articles consisting of 994 positive and 994 negative instances. Similarly to the PDTB-based data set, each positive instance represents a pair of articles at the same complexity level (i.e. either both complex or both simple). On the other hand, for each negative instance, we chose a pair of aligned articles from the SEW corpus (i.e. a pair of aligned articles containing one article taken from Wikipedia and its simpler version taken from SEW).

\subsection{Features for Predicting Text Complexity}
To predict text complexity, we have considered 16 individual features grouped into five classes. These are summarized in Table \ref{Features} and described below.

\begin{table*}
\centering
\small
\begin{tabular}{||l|l|l||}
\hline 
\textbf{Class of Features}& \textbf{Index} & \textbf{Feature Set}\\\hline\hline

Coherence features &\textit{F1} &  Log\_score of $<$realization-discourse relation$>$ \\
 &\textit{F2} & Log\_score of $<$discourse relation-discourse marker$>$ \\
 & \textit{F3}&Log\_score of $<$realization-discourse relation-discourse marker$>$\\
 &\textit{F4}& Discourse relation frequency \\ 
  \cdashline{1-3}

Cohesion features &\textit{F5}& Average \# of pronouns per sentence \\
 & \textit{F6}&Average \# of definite articles per sentence\\
 \cdashline{1-3}

Surface features & \textit{F7}&Text length \\
& \textit{F8} &Average \# of characters per word \\
& \textit{F9} & Average \# of words per sentence \\
\cdashline{1-3}

Lexical features & \textit{F10} &Average \# of word overlaps per sentence\\
& \textit{F11}& Average \# of synonyms of words in WordNet\\
& \textit{F12} & Average \# of frequency of words in Google Ngram corpus\\

 \cdashline{1-3}
 
Syntactic features & \textit{F13} &Average \# of verb phrases per sentence \\
 & \textit{F14}&Average \# of noun phrases per sentence \\
  & \textit{F15}&Average \# of subordinate clauses per sentence\\
  & \textit{F16}&Average height of syntactic parse tree \\ \hline

\end{tabular}
\caption{\small List of features in each class.}
\label{Features}
\end{table*}

\subsubsection{Coherence Features}
For a well written text to be coherent, utterances need to be connected logically and semantically using discourse relations. We considered coherence features in order to measure the association between this class of features and text complexity levels.  Our coherence features include: 

\medskip \textit{F1. }Pairs of $<$realization, discourse relations$>$ (e.g. \textit{$<$explicit, contrast$>$}).

\medskip \textit{F2. }Pairs of $<$discourse relations, discourse markers$>$, where applicable (e.g. \textit{$<$contrast, but$>$}).

\medskip \textit{F3. }Triplets of $<$discourse relations, realizations, discourse markers$>$, where applicable (e.g. \textit{$<$contrast, explicit, but$>$}).

\medskip \textit{F4. } Frequency of discourse relations.\\


Each article was considered as a bag of discourse properties. Then for features \textit{F1}, \textit{F2} and \textit{F3}, the log score of the probability of each article is calculated using Formulas (\ref{logLikelihood}) and (\ref{prob}). Considering a particular discourse feature (e.g. pairs of $<$discourse relations, discourse markers$>$), each article may contain a combination of $n$ occurrences of this feature with $k$ different feature values. The probability of observing such article is calculated using the multinomial probability mass function as shown in Formula (\ref{prob}). In order prevent arithmetic underflow and be more computationally efficient, we used the log likelihood of this probability mass function as shown in Formula (\ref{logLikelihood}).

\begin{gather}
\label{logLikelihood}
\begin{split}
log\_score(P) = log(P(n))+log(n!) + \\ \sum_{i=1}^{k} (x_i log(p_i)-log(x_i!))
\end{split}\\[2ex]
P = P(n) \dfrac{n!}{x_1!...x_k!} P_1...P_k
\label{prob}
\end{gather}

$P(n)$ is the probability of an article with $n$ instances of the feature we are considering, $x_i$ is the number of times a feature has its $i^{th}$ value and $P_i$ is the probability of a feature to have its $i^{th}$ value based on all the articles of the PDTB. For example, for the feature \textit{F1} (i.e. pair of $<$realization, discourse relation$>$), consider an article containing $<$\textit{explicit, contrast}$>$, $<$\textit{implicit, causality}$>$ and $<$\textit{explicit, contrast}$>$. In this case, $n$ is the total number of \textit{F1} features we have in the article (i.e. $n = 3$), and $P(n)$ is the probability of an article to have 3 such features across all PDTB articles. In addition, $x_1 = 2$ because we have two $<$\textit{explicit, contrast}$>$ pairs and $P_1$ is the probability of observing the pair $<$\textit{explicit, contrast}$>$ over all possible pairs of $<$realization, discourse relation$>$. Similarly, $x_2 = 1$ and $P_2$ is the probability of observing $<$\textit{implicit, causality}$>$ pair over all possible pairs of $<$realization, discourse relation$>$.

\subsubsection{Cohesion Features}
Cohesion is an important property of well-written texts \cite{grosz1995centering,barzilay2008}. Addressing an entity for the first time in a text is different from further mentions to the entity. Proper use of referencing influences the ease of following a text and subsequently its complexity. Pronoun resolution can affect text cohesion in the way that it prevents repetition. Also, according to \newcite{halliday1976cohesion}, definite description is an important characteristic of well-written texts. Thus, in order to measure the influence of cohesion on text complexity, we considered the following cohesive devices.

\medskip \textit{F5. }Average number of pronouns per sentence.

\medskip \textit{F6. }Average number of definite articles per sentence.

\subsubsection{Surface Features}
Surface features have traditionally been used in readability measures such as \cite{kincaid1975} to measure readability level. 
\newcite{pitler2009} showed that the only significant surface feature correlated with text complexity level was the length of the text. As a consequence, we investigated the influence of surface features by considering the following three surface features:

\medskip \textit{F7. }Text length as measured by the number of words.

\medskip \textit{F8. }Average number of characters per word.

\medskip \textit{F9. }Average number of words per sentence.

\subsubsection{Lexical Features}
In order to capture the influence of lexical choices across complexity levels, we considered the following three lexical features:

\medskip \textit{F10. }Average number of word overlaps per sentence.

\medskip \textit{F11. }Average number of synonyms of words in WordNet.

\medskip \textit{F12. }Average frequency of words in the Google N-gram (Web1T) corpus.
\vspace{0.3cm}

The lexical complexity of a text can be influenced by the number of words that are used in consecutive sentences. This means that if some words are used repetitively rather than introducing new words in the following sentences, the text should be simpler. This is captured by feature \textit{F10}: ``\textit{Average \# of word overlaps per sentence}'' which calculates the average number of word overlaps in all consecutive sentences. 

In addition, the number of synonyms of a word can be correlated to its complexity level. To account for this feature, \textit{F11}: ``Average \# of synonyms of words in WordNet'' is introduced to capture the complexity of the words \cite{miller1995wordnet}. Moreover, the frequency of a word can be an indicator of its simplicity. Also, feature \textit{F12}: ``\textit{Average \# of frequency of words in Google N-gram corpus}'' is used based on the assumption that simpler words are more frequently used. In order to measure the frequency of each word, we used the Google N-gram corpus \cite{michel2011}. Thus, pairs of articles at the same complexity level tend to have similar lexical features compared to pairs of articles at different complexity levels.

\begin{table*}
\centering
\small
\begin{tabular}{||l|c||C{2cm}|c|c||C{2cm}|c|c||}
\hline 


\textbf{Feature set} &  \textbf{No. features} &\textbf{SEW-based} & \textbf{p-value} & \textbf{Stat. Sign} & \textbf{PDTB} & \textbf{p-value} & \textbf{Stat. Sign}\\
 & & \textbf{ Data Set} & & &\textbf{ Data Set} & &\\
\hline\hline
Baseline &N/A &50.00\% & N/A & N/A & 51.23\% & N/A & N/A \\
All features & 16 &\bf94.96\% & N/A & N/A & \bf69.04\% & N/A & N/A\\\cdashline{1-3}
\cdashline{1-8}
Coherence only&4 &\bf93.76\% & 0.15 &$=$ & \bf64.02\% & 0.45 & $=$ \\
Cohesion only& 2&66.09\% & 0.00 & $\Downarrow$& 57.93\% & 0.01 & $\Downarrow$\\
Surface only& 3&83.45\% & 0.00 & $\Downarrow$ & 51.32\% & 0.00 & $\Downarrow$\\
Lexical only& 3&78.20\% & 0.00 & $\Downarrow$ &  46.29\% & 0.00 & $\Downarrow$ \\
Syntactic only& 4&79.32\% & 0.00 & $\Downarrow$ & 62.16\% & 0.24 & $=$\\
\cdashline{1-8}
All-Coherence & 12&\bf86.70\% & 0.00 & $\Downarrow$ & \bf62.43\% & 0.08 & $\Downarrow$\\
All-Cohesion & 14& 95.32\% & 0.44 & $=$ & 68.25\% & 0.76 & $=$\\
All-Surface & 13&95.10\% & 0.43 & $=$  & 68.25\% & 0.61 & $=$ \\
All-Lexical & 13&95.42\% & 0.38 & $=$ & 64.81\% & 0.57 & $=$\\
All-Syntactic &12& 94.30\% & 0.31 & $=$ & 66.40\% & 0.67 & $=$\\
\hline

\end{tabular}
\caption{\small Accuracy of Random Forest models built using different subset of features.}
\label{Results}
\end{table*}

\subsubsection{Syntactic Features}
According to \newcite{kate2010}, syntactic structures seem to affect text complexity level. As \newcite{barzilay2008} note, more noun phrases make texts more complex and harder to understand. In addition, \newcite{bailin2001} pointed out that the use of multiple verb phrases in a sentence can make the communicative goal of a text more clear as explicit discourse markers will be used to connect them; however it can also make a text harder to understand for less educated adults or children. The \newcite{schwarm2005} readability assessment model was built based on a trigram language model, syntactic and surface features. Based on these previous work, we used the same syntactic features which includes:

\medskip \textit{F13. }Average number of verb phrases per sentence.

\medskip \textit{F14. }Average number of noun phrases per sentence.

\medskip \textit{15. }Average number of subordinate clauses per sentence.

\medskip \textit{F16. }Average height of syntactic parse tree.\\

These features were determined using the Stanford parser \cite{toutanova2003}.

\subsection{Results and Analysis}
In order to investigate the influence of each class of feature to assess the complexity level of a given pair of articles, we built several Random Forest classifiers and experimented with various subsets of features. Table \ref{Results} shows the accuracy of the various classifiers on our data sets (see Section \ref{data}) using 10-fold cross-validation. In order to test the statistical significance of the results, we conducted a two-sample t-test (with a confidence level of 90\%) comparing the models built using each feature set to the model trained using all features. A statistically significant decrease ($\Downarrow$) or no difference ($=$) is specified in the column labeled Stat. Sign.   

Our baseline is to consider no feature and simply assign the class label of the majority class. As indicated in Table \ref{Results}, the baseline is about 50\% for both data sets. When all features are used, the accuracy of the classifier trained on the SEW-based data set is 94.96\% and the one trained on the PDTB-based data set is 69.04\%. 

Considering only one class of features, the models trained using coherence features on both data sets outperformed the others (93.76\% and 64.02\%) and their accuracy are statistically as high as using all features together. However one must also note that there is a significant difference between the number of features (4 for coherence only vs. 16 for all features). Indeed, in both data sets, cohesion features are more useful than lexical features and less than syntactic features. 

Furthermore, it is interesting to note that surface features seem to be more discriminating in the SEW articles rather than in PDTB articles; however, viceversa is true about cohesion features. In addition, the decrease in the accuracy of all classifiers trained on the SEW using only one feature except coherence features is statistically significant. The same is true about the models trained on the PDTB with the only difference being the one trained using only syntactic features which performs as well as the one trained using all the features (62.16\% vs. 69.04\%). 

The last section of Table \ref{Results} shows the classification results when excluding only one class of features. In this case, removing coherence features leads to a more significant drop in performance compared to the other classes of features. The classifier trained using all features except the coherence features achieves an accuracy of 86.70\% and 62.43\% on the SEW and PDTB corpus respectively. This decrease in both models is statistically significant; however the changes in the accuracy of other classifiers trained using all features excluding only one class is not statistically significant.

\subsection{Feature Selection}
In any classification problem, feature selection is useful to identify the most discriminating features and reduce the dimensionality and model complexity by removing the least discriminating ones. In this classification problem, we built several classifiers using different subsets of features; however, identifying how well a feature can discriminate the classes would be helpful in building a more efficient model with fewer number of features.

\begin{table*}
    \centering
    \small
    \begin{tabular}{||l|l||l|l||}
    \hline
    \multicolumn{1}{||c|}{\bf Index}&\multicolumn{1}{c||}{\bf SEW-based Data Set} &\multicolumn{1}{c|}{\bf Index} &\multicolumn{1}{c||}{\bf PDTB-based Data Set}\\ \hline
    
    \textit{F2} &Log\_score of $<$discourse relation-marker$>$& \textit{F1}& Log\_score of $<$realization-discourse relation$>$  \\
   
     \textit{F9}&Average \# of words per sentence & \textit{F3}&Log\_score of $<$realization-relation-marker$>$ \\
    \textit{F14}& Average \# of noun phrases per sentence & \textit{F4}&Discourse relation frequency  \\
    \textit{F7}&Text length & \textit{F5}&Average \# of pronouns per sentence  \\

    \textit{F16}&Average height of syntactic parse tree & \textit{F9}&Average \# of words per sentence \\
    \textit{F13}&Average \# of verb phrases per sentence & \textit{F2}& Log\_score of $<$discourse relation-marker$>$ \\
    \textit{F15}&Average \# of subordinate clauses per sentence & \textit{F7}&Text length \\
    \textit{F10}&Average \# of word overlaps per sentence & \textit{F8}&Average \# of characters per word \\ 
     \textit{F8}&Average \# of characters per word &\textit{F12} & Average frequency of words in Web1T corpus\\
    
     \textit{F4}&Discourse relation frequency & \textit{F11} & Average \# of synonyms of words in WordNet\\
     \textit{F6}&Average \# of definite articles per sentence & \textit{F6}&Average \# of definite articles per sentence\\
    \textit{F11} & Average \# of synonyms of words in WordNet & \textit{F10}&Average \# of word overlaps per sentence \\
    \textit{F3}&Log\_score of $<$realization-relation-marker$>$ & \textit{F15}& Average \# of subordinate clauses per sentence \\
    \textit{F1}&Log\_score of $<$realization-discourse relation$>$ & \textit{F14}&Average \# of noun phrases per sentence \\
    \textit{F12} & Average frequency of words in Web1T corpus & \textit{F13}&Average \# of verb phrases per sentence \\
    \textit{F5}&Average \# of pronouns per sentence & \textit{F16}&Average height of syntactic parse tree \\
    
    \hline
     
    \end{tabular}
    \caption{Features ranked by information gain}
    \label{featureSelection}
\end{table*}

Using our pairwise classifier built with all the features, we ranked the features by their information gain. Table \ref{featureSelection} shows all the features used in the two models using all the features trained on the PDTB-based data set and the SEW-based data set.

As can be seen in Table \ref{featureSelection}, coherence features are among the most discriminating features on the PDTB-based data set as they hold the top three positions. Also, the most discriminating feature on the SEW-based data set is a coherence feature. We investigated the power of only the top feature in both data sets by classifying the data using only this single feature and evaluated using 10-fold cross-validation. Using only \textit{F1}: ``\textit{log\_score of $<$realization, discourse relation$>$}'' to classify the PDTB-based data set, we achieved an accuracy of 56.34\%. This feature on its own outperformed the individual class of surface features and lexical features and performed as well as combining the features of the two classes (four features). It also performed almost as well as the two cohesion features (\textit{F5}, \textit{F6}). In addition, using only the feature \textit{F2}: ``\textit{log\_score of $<$discourse relation, discourse marker$>$}'' on the SEW corpus resulted in an accuracy of 77.26\% which is much higher than the accuracy of the classifier built using the class of cohesion and almost as good as lexical features.

\section{Conclusion}
In this paper we investigated the influence of various classes of features in pairwise text complexity assessment on two data sets created from standard corpora. The combination of 16 features, grouped into five classes of surface, lexical, syntactic, cohesion and coherence features resulted in the highest accuracy. However the use of only 4 coherence features performed statistically as well as using all features on both data sets. 

In addition, removing only one class of features from the combination of all the features did not affect the accuracy; except for coherence features. Removing the class of coherence features from the combination of all features led to a statistically significant decrease in accuracy. Thus, we can conclude a strong correlation between text coherence and text complexity. This correlation is weaker for other classes of features.

\subsection*{Acknowledgement}
The authors would like to thank the anonymous reviewers for their feedback on the paper. This work was financially supported by NSERC.



\bibliographystyle{acl2016}

\appendix

\end{document}